\documentclass[conference]{IEEEtran}

\usepackage[T1]{fontenc}
\usepackage[utf8]{inputenc}
\usepackage{cite}
\usepackage{amsmath,amssymb,amsfonts}
\usepackage{algorithmic}
\usepackage{graphicx}
\usepackage{textcomp}
\usepackage{xcolor}
\usepackage{booktabs}
\usepackage{array}
\usepackage{multirow}
\usepackage{url}
\usepackage{hyperref}
\usepackage{pifont}
\newcommand{\cmark}{\ding{51}}
\newcommand{\xmark}{\ding{55}}
\hypersetup{colorlinks=true, linkcolor=blue, citecolor=blue, urlcolor=blue}

\def\BibTeX{{\rm B\kern-.05em{\sc i\kern-.025em b}\kern-.08em
    T\kern-.1667em\lower.7ex\hbox{E}\kern-.125emX}}

\begin{document}

\title{MiniVLA-Nav v1: A Multi-Scene Simulation Dataset for\\
Language-Conditioned Robot Navigation}

\author{Ali Al-Bustami \and Jaerock Kwon%
\thanks{Department of Electrical and Computer Engineering,
University of Michigan-Dearborn, Dearborn, USA.
Email: \{abustami, jrkwon\}@umich.edu.}}

\IEEEoverridecommandlockouts
\maketitle

\begin{abstract}
We present \textbf{MiniVLA-Nav v1}, a simulation dataset for
\emph{Language-Conditioned Object Approach} (LCOA) navigation:
given a short natural-language instruction, an NVIDIA Nova Carter
differential-drive robot must navigate to the named object and stop
within 1\,m across four photorealistic Isaac Sim environments (Office,
Hospital, Full Warehouse, and Warehouse with Multiple Shelves).
Each of the 1\,174 episodes pairs an
instruction with synchronized 640\texttimes{}640 RGB images, metric
depth maps (float32, metres), and instance segmentation masks, together
with continuous $(v,\omega)$ and 7\texttimes{}7 tokenized expert action
labels recorded at 60\,Hz from a vision-based proportional controller.
Trajectory diversity is ensured through three spawn-distance tiers
(near: 1.5--3.5\,m, mid: 3.5--7.0\,m, far: global curated points;
Pearson $r = 0.94$ between spawn distance and trajectory length),
12 object categories, 18 training templates, and 12 paraphrase-OOD templates.
Five evaluation splits support in-distribution accuracy,
template-paraphrase robustness, and OOD object-category benchmarking.
The dataset is publicly available at
\url{https://huggingface.co/datasets/alibustami/miniVLA-Nav}.
\end{abstract}


\section{Introduction}
\label{sec:intro}

Teaching robots to follow natural-language navigation instructions is a
foundational challenge in embodied AI. Recent vision-language-action (VLA)
models \cite{brohan2023rt2, kim2024openvla, octo2024} demonstrate that
language-conditioned policies can be trained end-to-end from large demonstration
datasets, yet acquiring diverse, high-quality robot demonstrations at scale
remains expensive and logistically difficult in the real world.

Simulation offers a practical remedy: physics-faithful environments generate
demonstrations at zero marginal cost with perfect ground-truth annotation.
However, existing datasets have key gaps: continuous-action navigation datasets
with dense per-step expert labels remain rare
\cite{krantz2020r2rce}; large-scale object-goal navigation benchmarks
use discrete panoramic actions rather than differential-drive control
\cite{chaplot2020object}; and language-annotated demonstration datasets
target tabletop manipulation rather than wheeled navigation
\cite{shridhar2020alfred}.
Critically, no publicly available dataset combines \emph{multiple indoor scene
types}, \emph{continuous differential-drive expert actions}, \emph{language
instructions}, and \emph{multi-modal per-step observations} (RGB + depth +
segmentation) in a single collection.

This paper introduces \textbf{MiniVLA-Nav v1}, a dataset designed for the
\emph{Language-Conditioned Object Approach} (LCOA) task: given a short
natural-language instruction and front RGB-D observations, a differential-drive
robot must navigate to the named object and stop within 1\,m.
Our primary contributions are:

\begin{enumerate}
\item A \textbf{scalable, resumable data-generation pipeline} built on
      NVIDIA Isaac Sim 5.1, automating scene loading, tiered spawn
      sampling, expert rollout, and structured episode archiving with
      full reproducibility from a single random seed.

\item \textbf{Four indoor scenes} spanning office, hospital, and warehouse
      domains, with 12 object categories, three spawn-distance tiers, and
      per-scene seen/held-out category configurations.

\item \textbf{Multi-modal per-timestep observations}: synchronized
      640\texttimes{}640 RGB images, metric depth maps (float32, metres),
      and instance segmentation masks, paired with continuous $(v,\omega)$
      and 7\texttimes{}7 discretized action tokens.

\item \textbf{Five evaluation splits} supporting in-distribution accuracy,
      template-paraphrase OOD robustness, and OOD object-category
      generalization, with curated heldout categories per scene.

\item Comprehensive \textbf{dataset statistics} characterizing trajectory
      length, spawn distance, tier difficulty, instruction template
      coverage, and object frequency across 1,174 validated episodes.
\end{enumerate}

This paper releases the dataset, generation pipeline, and evaluation
tooling; training baselines will appear in a companion publication.

\section{Related Work}
\label{sec:related}

\subsection{Vision-and-Language Navigation}

Anderson \emph{et al.}~\cite{anderson2018r2r} introduced Room-to-Room (R2R),
requiring agents to follow multi-step instructions in panoramic indoor
environments; follow-up work extended this to continuous action spaces
\cite{krantz2020r2rce} and outdoor scenes \cite{chen2019touchdown}.
Gu \emph{et al.}~\cite{gu2022vlnsurvey} survey the full VLN landscape and
identify single-step approach tasks with dense action labels as an
under-explored regime — the gap MiniVLA-Nav directly targets.
MiniVLA-Nav is complementary to multi-step VLN: it focuses on the
\emph{approach} sub-problem with continuous differential-drive control
and dense per-step imitation labels.

\subsection{Simulated Robot Datasets}

The Habitat platform \cite{habitat2021} provides large-scale rearrangement
and navigation benchmarks with discrete panoramic actions; continuous-action
extensions \cite{krantz2020r2rce} exist but lack language conditioning
and multi-scene diversity.
The ObjectNav benchmark \cite{batra2020objectnav} targets object-goal
finding using discrete panoramic actions without language instructions;
Chaplot et al.~\cite{chaplot2020object} extend it with semantic exploration
but still use discrete actions.
ALFRED \cite{shridhar2020alfred} provides language-annotated demonstrations
but targets tabletop manipulation in static scenes, not mobile navigation.
RoboVQA \cite{sermanet2023robovqa} and OpenX \cite{open_x_embodiment_rt_x_2023}
aggregate large real-robot datasets but lack systematic OOD evaluation splits
and are focused on manipulation rather than wheeled navigation.
MiniVLA-Nav v1 addresses the specific gap of multi-scene, continuous-action,
language-conditioned wheeled-navigation demonstrations with structured
OOD evaluation.

\subsection{Language-Conditioned Control}

RT-2 \cite{brohan2023rt2} and OpenVLA \cite{kim2024openvla} show that
VLA models fine-tuned on imitation data can follow novel instructions
zero-shot, but both target tabletop manipulation.
To our knowledge, MiniVLA-Nav v1 is the first dataset combining
multiple indoor scene types, continuous differential-drive expert trajectories,
multi-modal per-step observations, and language conditioning with systematic
OOD evaluation splits.

Table~\ref{tab:related} positions MiniVLA-Nav v1 against representative
prior datasets on key dimensions.

\begin{table}[t]
\caption{Comparison with Related Datasets}
\label{tab:related}
\centering
\renewcommand{\arraystretch}{1.1}
\resizebox{\columnwidth}{!}{%
\begin{tabular}{@{}lccccc@{}}
\toprule
\textbf{Dataset} & \textbf{Sim?} & \textbf{Action} & \textbf{Lang.} & \textbf{Scenes} & \textbf{OOD splits} \\
\midrule
R2R \cite{anderson2018r2r}          & \xmark & Discrete  & \cmark & 90  & \xmark \\
VLN-CE \cite{krantz2020r2rce}       & \cmark & Cont.     & \cmark & 1   & \xmark \\
ObjectNav \cite{batra2020objectnav} & \cmark & Discrete  & \xmark & 1   & \xmark \\
ALFRED \cite{shridhar2020alfred}    & \cmark & Discrete  & \cmark & 1   & \xmark \\
\textbf{MiniVLA-Nav v1 (ours)}      & \cmark & \textbf{Cont.} & \cmark & \textbf{4} & \cmark \\
\bottomrule
\end{tabular}}
\end{table}

\section{Task Definition}
\label{sec:task}

\subsection{Language-Conditioned Object Approach (LCOA)}

Given a natural-language instruction $\ell$ and a stream of front-facing
observations $o_t = (\mathbf{I}_t^{\text{RGB}}, \mathbf{D}_t)$, the robot
must output a sequence of actions $a_t = (v_t, \omega_t)$ such that
\begin{equation}
  \|p_T - p_g\| \leq r_{\text{success}} = 1.0\,\text{m},
\end{equation}
where $p_T$ is the robot position at termination and $p_g$ is the
3-D centroid of the target object bounding box.

\textbf{Episode termination} occurs on one of three conditions:
\begin{itemize}
  \item \emph{Success}: robot within $r_{\text{success}}$ and stationary
        for $\geq 5$ consecutive steps (stopped-hold criterion).
  \item \emph{Collision}: robot commanded forward but makes no progress
        for $\geq 16$ consecutive steps with a near obstacle
        (stall detection).
  \item \emph{Timeout}: maximum $T_{\max} = 1000$ steps reached without success.
\end{itemize}

Only successful episodes are retained in the released dataset.

\subsection{Action Space}

\begin{figure}[t]
  \centering
  \includegraphics[width=0.9\columnwidth]{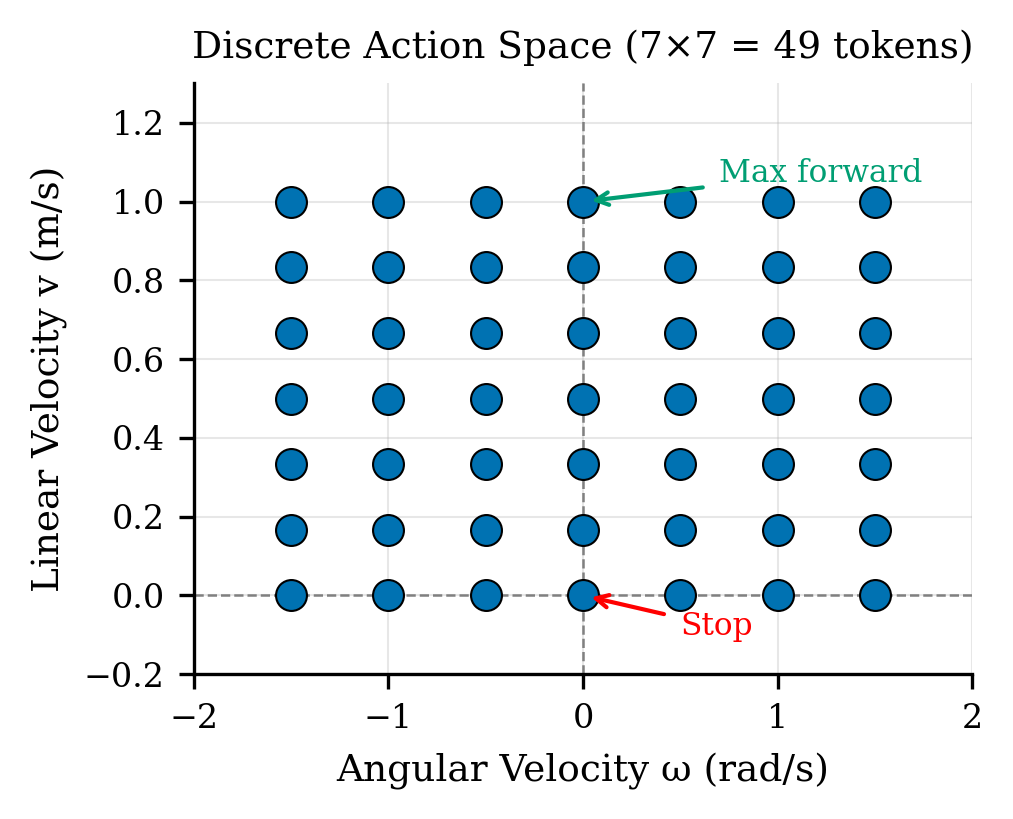}
  \caption{Discrete 7\texttimes{}7 action space.
           Each grid point is one valid ($v$, $\omega$) token.
           The origin (stop) and maximum-forward tokens are annotated.}
  \label{fig:action_space}
\end{figure}

The continuous action is $(v, \omega) \in [0, 1]\,\text{m/s} \times [-1.5, 1.5]\,\text{rad/s}$.
For VLA-style token prediction, each dimension is quantized to 7 uniform
bins, yielding a 7\texttimes{}7 = 49-token joint action vocabulary
(Fig.~\ref{fig:action_space}).

\section{Data Collection Pipeline}
\label{sec:pipeline}

\subsection{Simulation Environment}

All data were collected in NVIDIA Isaac Sim 5.1
(release \texttt{5.1.0-rc.19}), which provides GPU-accelerated
rigid-body physics (PhysX), photorealistic rendering (RTX),
and a Python API for programmatic scene control.
The physics time step is $\Delta t = 1/60\,\text{s}$, so the expert
controller and sensor observations operate at \textbf{60\,Hz}.

\subsection{Robot Platform}

We use the \textbf{NVIDIA Nova Carter}, a differential-drive platform
with a 0.52\,m wheelbase and 0.14\,m wheel radius.
The onboard front-facing stereo camera (\texttt{front\_hawk/right})
is configured at 640\texttimes{}640 pixels and provides synchronized RGB,
distance-to-image-plane depth (float32, metres), and instance
segmentation frames at every simulation step.

\subsection{Scene Catalog and Target Discovery}

Four scene USDs from the Isaac Assets library are used
(Table~\ref{tab:scenes}).
At the first episode of each scene, the USD stage is traversed to
discover navigable targets by matching prim names against 12 category
rules (\texttt{chair}, \texttt{sofa}, \texttt{table}, \texttt{monitor},
\texttt{plant}, \texttt{trash\_can}, \texttt{fire\_extinguisher},
\texttt{whiteboard}, \texttt{shelf}, \texttt{rack}, \texttt{barrel},
\texttt{crate}).
Discovered targets and their 3-D bounding-box centroids are cached in
per-scene \texttt{targets\_<scene>.yaml} files.

\begin{table}[t]
\caption{Scene Configuration}
\label{tab:scenes}
\centering
\renewcommand{\arraystretch}{1.15}
\resizebox{\columnwidth}{!}{%
\begin{tabular}{@{}lcc@{}}
\toprule
\textbf{Scene} & \textbf{Seen cats.} & \textbf{Held-out cats.} \\
\midrule
Office                  & chair, sofa, table, monitor, plant, trash\_can & fire\_ext., whiteboard \\
Hospital                & chair, trash\_can                              & fire\_ext., whiteboard \\
Full Warehouse          & shelf, rack                                    & barrel, crate \\
Warehouse (Multi-Shelf) & shelf, rack                                    & barrel, crate \\
\midrule
\multicolumn{3}{c}{12 categories total} \\
\bottomrule
\end{tabular}}
\end{table}

\subsection{Tiered Spawn Sampling}
\label{subsec:spawn}

\begin{figure}[t]
  \centering
  \includegraphics[width=\columnwidth]{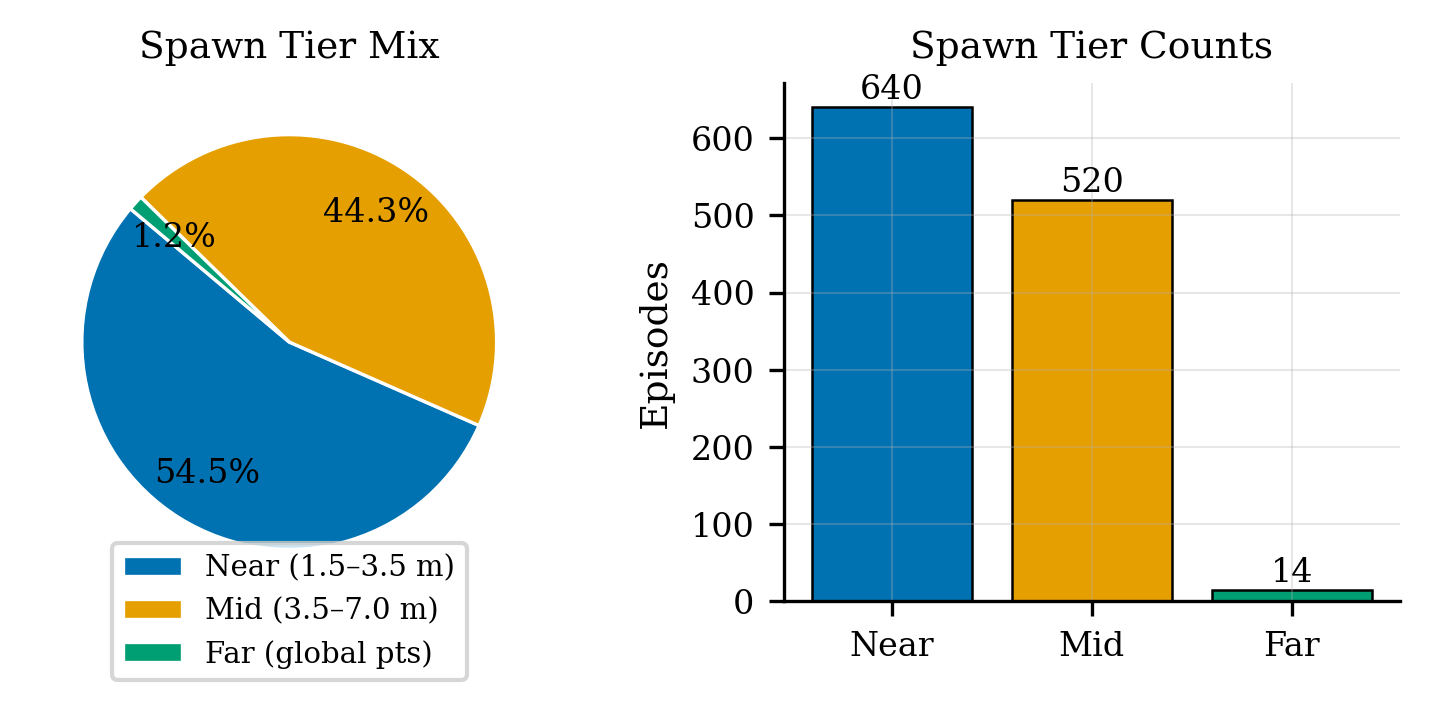}
  \caption{Spawn-tier distribution. The current 1\,174-episode snapshot
           contains 640 near-tier (54.5\%), 520 mid-tier (44.3\%),
           and 14 far-tier (1.2\%) episodes.
           Far-tier remains under-represented because global spawn-point files
           cover only a subset of scenes and floor positions.}
  \label{fig:spawn_tiers}
\end{figure}

A key design choice in v1 is \emph{tiered spawn sampling}
to ensure diversity in navigation difficulty
(Fig.~\ref{fig:spawn_tiers}).
Each episode independently samples one of three tiers:

\begin{itemize}
  \item \textbf{Near} (30\%): spawn uniformly at distance $r \sim \mathcal{U}(1.5, 3.5)$\,m
        from the target, facing roughly toward it with $\pm 25^{\circ}$ heading noise.
  \item \textbf{Mid} (40\%): same procedure with $r \sim \mathcal{U}(3.5, 7.0)$\,m.
  \item \textbf{Far} (30\%): sample from a precomputed set of global valid floor positions
        (validated via displacement check after physics warmup), with uniform random heading.
\end{itemize}

After placement, spawn validity is confirmed by measuring robot displacement
from the intended position after three warmup steps; positions with
displacement $>$ 0.08\,m are retried up to 8 times.

\subsection{Language Instruction Generation}

Instructions are generated by filling slot templates with target
category names and optional color attributes.
Table~\ref{tab:templates} lists the 18 training and 12 OOD templates.
At each episode, a template is sampled uniformly from the pool
appropriate to the episode's split; color-slot templates (those
containing \texttt{\{color\}}) are excluded when the target has no
color annotation.
In the current v1 release, \emph{all} targets carry
\texttt{color\,=\,"unknown"} because the USD assets do not expose
material-color attributes through a standard prim API; consequently,
color-slot templates are suppressed for every episode and Fig.~\ref{fig:templates}
reflects the usage of non-color templates only.
A mesh-based dominant-color extraction step is planned for v1.1.

\begin{table}[t]
\caption{Complete Language Template Inventory}
\label{tab:templates}
\centering
\renewcommand{\arraystretch}{1.05}
\resizebox{\columnwidth}{!}{%
\begin{tabular}{@{}clp{5.8cm}@{}}
\toprule
\textbf{ID} & \textbf{Pool} & \textbf{Template} \\
\midrule
T1  & Train & ``Go to the \{object\}.'' \\
T2  & Train & ``Drive to the \{object\} and stop.'' \\
T3  & Train & ``Approach the \{object\}.'' \\
T4  & Train & ``Move toward the \{object\}.'' \\
T5  & Train & ``Navigate to the \{object\}.'' \\
T6  & Train & ``Go to the \{color\} \{object\}.''$^\dagger$ \\
T7  & Train & ``Drive to the \{color\} \{object\} and stop.''$^\dagger$ \\
T8  & Train & ``Approach the \{color\} \{object\}.''$^\dagger$ \\
T9  & Train & ``Head to the \{object\}.'' \\
T10 & Train & ``Move to the \{object\} and halt.'' \\
T11 & Train & ``Go to the \{object\} in front of you.'' \\
T12 & Train & ``Drive toward the \{object\} until you reach it.'' \\
T13 & Train & ``Get to the \{object\}.'' \\
T14 & Train & ``Your destination is the \{object\}.'' \\
T15 & Train & ``Locate the \{object\} and stop in front of it.'' \\
T16 & Train & ``Go over to the \{color\} \{object\}.''$^\dagger$ \\
T17 & Train & ``Move all the way to the \{object\}.'' \\
T18 & Train & ``Stop next to the \{object\}.'' \\
\midrule
O1  & Paraphrase-OOD & ``Make your way to the \{object\}.'' \\
O2  & Paraphrase-OOD & ``Proceed to the \{object\}.'' \\
O3  & Paraphrase-OOD & ``Find the \{object\} and come to a stop.'' \\
O4  & Paraphrase-OOD & ``Roll over to the \{color\} \{object\}.''$^\dagger$ \\
O5  & Paraphrase-OOD & ``Go straight to the \{object\}.'' \\
O6  & Paraphrase-OOD & ``Close in on the \{object\}.'' \\
O7  & Paraphrase-OOD & ``Reach the \{object\} and stop there.'' \\
O8  & Paraphrase-OOD & ``Move until you're beside the \{object\}.'' \\
O9  & Paraphrase-OOD & ``Find your way to the \{object\}.'' \\
O10 & Paraphrase-OOD & ``Head toward the \{color\} \{object\} and stop.''$^\dagger$ \\
O11 & Paraphrase-OOD & ``Get closer to the \{object\}.'' \\
O12 & Paraphrase-OOD & ``Park next to the \{object\}.'' \\
\bottomrule
\end{tabular}}
\vspace{2pt}
{\small $^\dagger$Color-slot templates; suppressed in v1 (all targets have \texttt{color=unknown}).
Active pool: 13 train + 10 paraphrase-OOD.}
\end{table}

\subsection{Expert Controller}

The expert is a \emph{proportional controller} that uses pixel-level
target visibility from the instance segmentation mask.
When the target subtends $\geq 32$ pixels:
\begin{equation}
\omega_t = \text{clamp}\!\left(-k_\omega\,\frac{c_x - W/2}{W/2}, \,
                               \omega_{\min}, \omega_{\max}\right),
\end{equation}
\begin{equation}
v_t = \text{clamp}\!\left(k_v\,(\hat{d}_t - r_{\text{success}}), \,
                           0, v_{\max}\right),
\end{equation}
where $c_x$ is the target mask centroid column, $W = 640$,
$\hat{d}_t$ is the \emph{median} depth (metres) sampled over all target
mask pixels (a robust estimate of object distance),
$k_\omega = 1.4$\,rad$^{-1}$, and $k_v = 0.7$\,s$^{-1}$.
When the target is not visible, the controller uses a bearing-only
proportional law computed from the known goal position:
\begin{equation}
\omega_t = \text{clamp}\!\left(k_\omega\,\Delta\psi_t,\, \omega_{\min}, \omega_{\max}\right),
\end{equation}
\begin{equation}
v_t = \text{clamp}\!\left(k_v\,d_t,\, 0, v_{\max}\right),
\end{equation}
where $\Delta\psi_t$ is the heading error and $d_t$ is the Euclidean
distance to the goal.
An obstacle-avoidance layer clamps $v$ when depth in the central
foreground crop (rows 55--95\%, cols 25--75\%) falls below 0.25\,m.

\subsection{Episode Archiving}

\begin{figure*}[t]
  \centering
  \includegraphics[width=\textwidth]{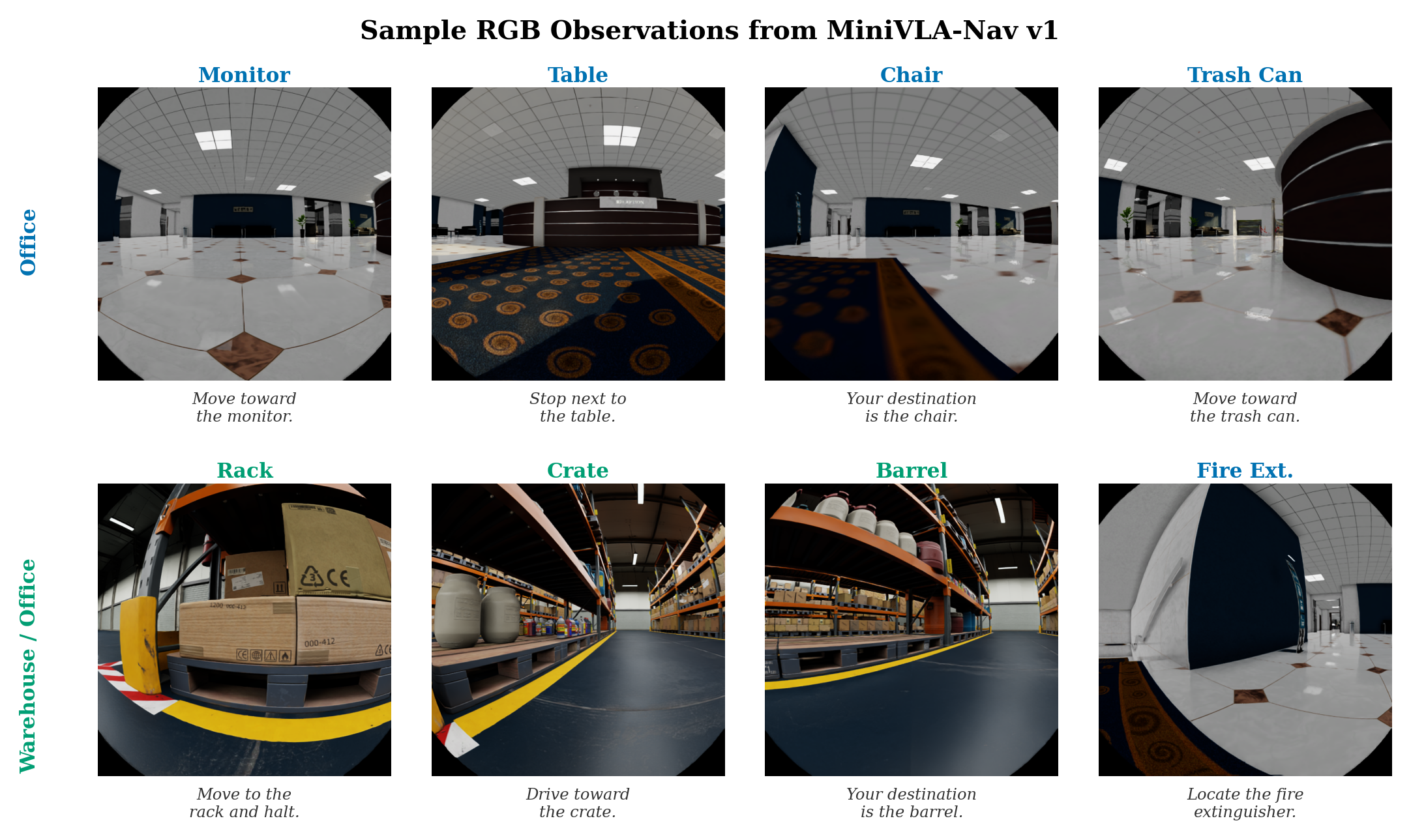}
  \caption{Sample front-facing RGB observations from MiniVLA-Nav v1.
           Each panel shows a representative frame with the target category
           (bold, colour-coded by scene) and the episode instruction (italic).
           \textbf{Office} (blue, top row): monitor, table, chair, trash can.
           \textbf{Bottom row}: rack, crate, barrel (Full Warehouse) and
           fire extinguisher (Office, held-out category).
           All images are 640\texttimes{}640 pixels captured by the Nova Carter
           front stereo camera under Isaac Sim RTX rendering.}
  \label{fig:samples}
\end{figure*}

Each successful episode is stored in a self-contained directory
\texttt{episodes/ep\_\{N:06d\}/} with the structure in
Table~\ref{tab:episode_struct}.
Fig.~\ref{fig:samples} shows representative RGB frames from all four
scenes across eight object categories.
A \texttt{meta.json} sidecar records the full episode configuration:
scene path, robot and camera configuration, target prim path,
goal centroid, instruction text and template ID, spawn tier and
distance, rollout statistics, and ISO-8601 UTC timestamp.

\begin{table}[t]
\caption{Episode Directory Structure}
\label{tab:episode_struct}
\centering
\renewcommand{\arraystretch}{1.1}
\resizebox{\columnwidth}{!}{%
\begin{tabular}{@{}lll@{}}
\toprule
\textbf{File} & \textbf{Shape / Format} & \textbf{Description} \\
\midrule
\texttt{rgb\_front/\{t\}.png}    & $640\!\times\!640\!\times\!3$, uint8  & Front RGB \\
\texttt{depth\_front/\{t\}.npy}  & $640\!\times\!640$, float32           & Depth (m) \\
\texttt{seg\_front/\{t\}.png}    & $640\!\times\!640$, uint16            & Instance seg. \\
\texttt{actions\_continuous.npy} & $T\!\times\!2$, float32               & $(v_t, \omega_t)$ \\
\texttt{actions\_tokens.npy}     & $T\!\times\!2$, int16                 & Discretized tokens \\
\texttt{poses.npy}               & $T\!\times\!7$, float32               & $(x,y,z,q_w,q_x,q_y,q_z)$ \\
\texttt{meta.json}               & JSON                                  & Episode metadata \\
\bottomrule
\end{tabular}}
\end{table}

\section{Dataset Statistics}
\label{sec:stats}

\subsection{Collection Status}

Table~\ref{tab:overview} summarises the collection status.
A total of 1\,174 episodes have been validated across all four scenes,
with the Office scene fully complete at its 700-episode budget.

\textbf{Expert success rate.}
Only successful episodes are retained; failed attempts are discarded.
The proportional expert's attempt-to-success rate varies by scene:
the open-plan Office achieves $\sim$80--90\%, while the Full Warehouse
and Hospital yield $\sim$40--60\% due to narrow corridors triggering
stall detection.
A pre-generation expert acceptance gate (pilot run of 50 episodes per
scene, requiring $\geq$80\% SR before full generation) was applied and
passed for the Office scene; the warehouse and hospital scenes proceeded
on a best-effort basis given the controller's structural limitations.
These rates introduce a selection bias toward unobstructed approach paths;
future work will address this with a path-planning expert.

\begin{figure}[t]
  \centering
  \includegraphics[width=\columnwidth]{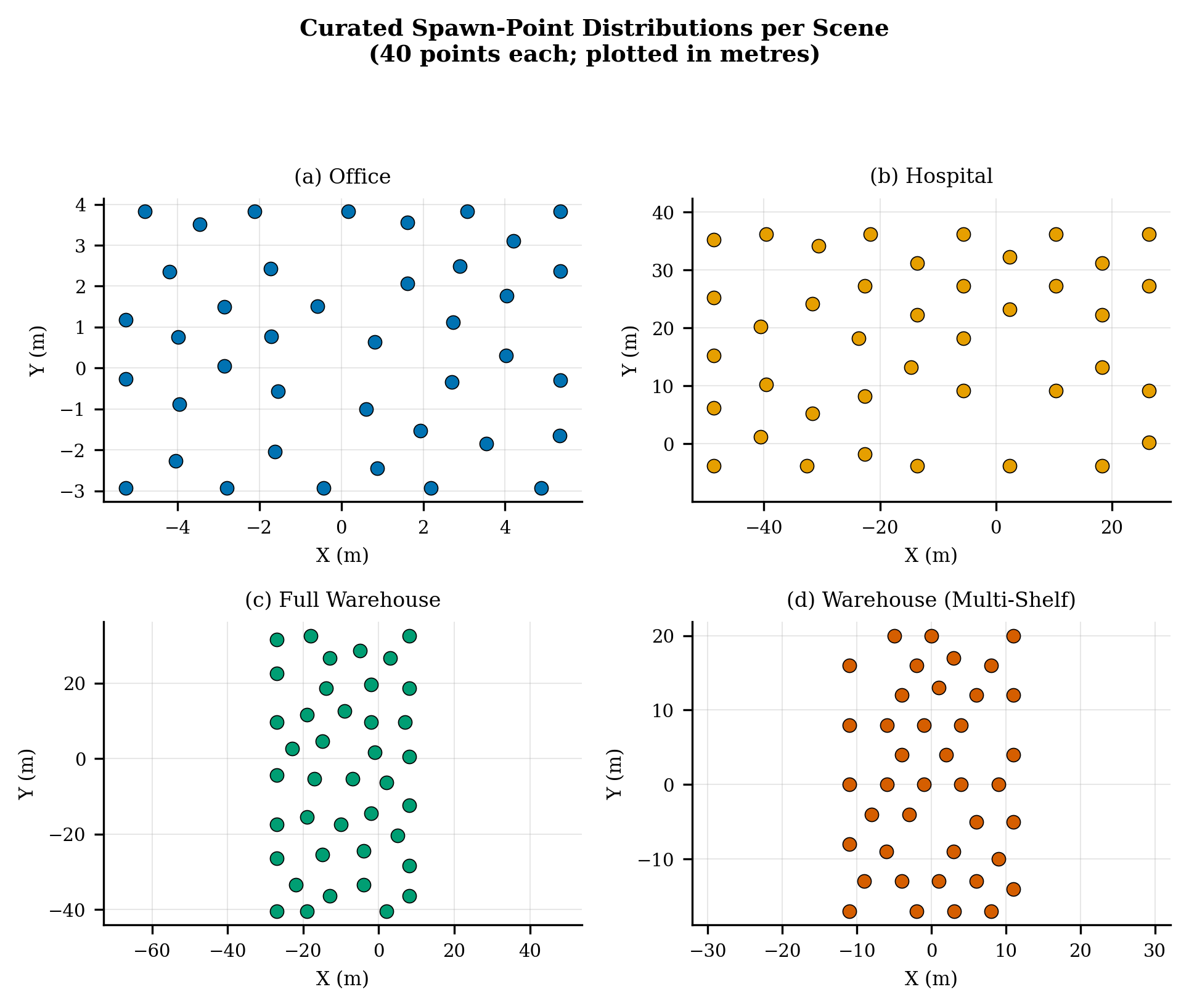}
  \caption{Curated spawn-point distributions (40 points each) per scene.
           Each point marks a validated floor position confirmed
           free of collision after a 3-step physics warmup.
           The Office scene uses cm-scale USD units (values converted to
           metres for display); other scenes use metre units directly.}
  \label{fig:spawn_maps}
\end{figure}

Fig.~\ref{fig:spawn_maps} shows the curated spawn-point distributions
for each scene.

\begin{table}[t]
\caption{Episodes per Scene}
\label{tab:overview}
\centering
\renewcommand{\arraystretch}{1.15}
\begin{tabular}{@{}lrr@{}}
\toprule
\textbf{Scene}            & \textbf{Episodes} & \textbf{Mean TL (m)} \\
\midrule
Office                    & 700  & 2.66 \\
Hospital                  &  52  & 2.73 \\
Full Warehouse            & 354  & 2.83 \\
Warehouse (Multi-Shelf)   &  68  & 3.15 \\
\midrule
\textbf{Total}            & \textbf{1174} & \textbf{2.74} \\
\bottomrule
\end{tabular}
\end{table}

\subsection{Split Distribution}

Episodes are assigned to one of five splits with ratios
60\,/\,10\,/\,10\,/\,10\,/\,10 per scene:

\begin{itemize}
  \item \texttt{train\_id} (716): seen objects, seen templates.
  \item \texttt{val\_id}   (114): seen objects, seen templates (validation).
  \item \texttt{test\_id}  (121): seen objects, seen templates (held-out test).
  \item \texttt{test\_paraphrase\_ood} (122): seen objects, \emph{paraphrase templates}.
        These are syntactic reformulations of training templates, not
        semantically out-of-distribution instructions.
  \item \texttt{test\_ood\_obj}  (101): \emph{heldout object categories}, seen templates.
\end{itemize}

\begin{figure}[t]
  \centering
  \includegraphics[width=\columnwidth]{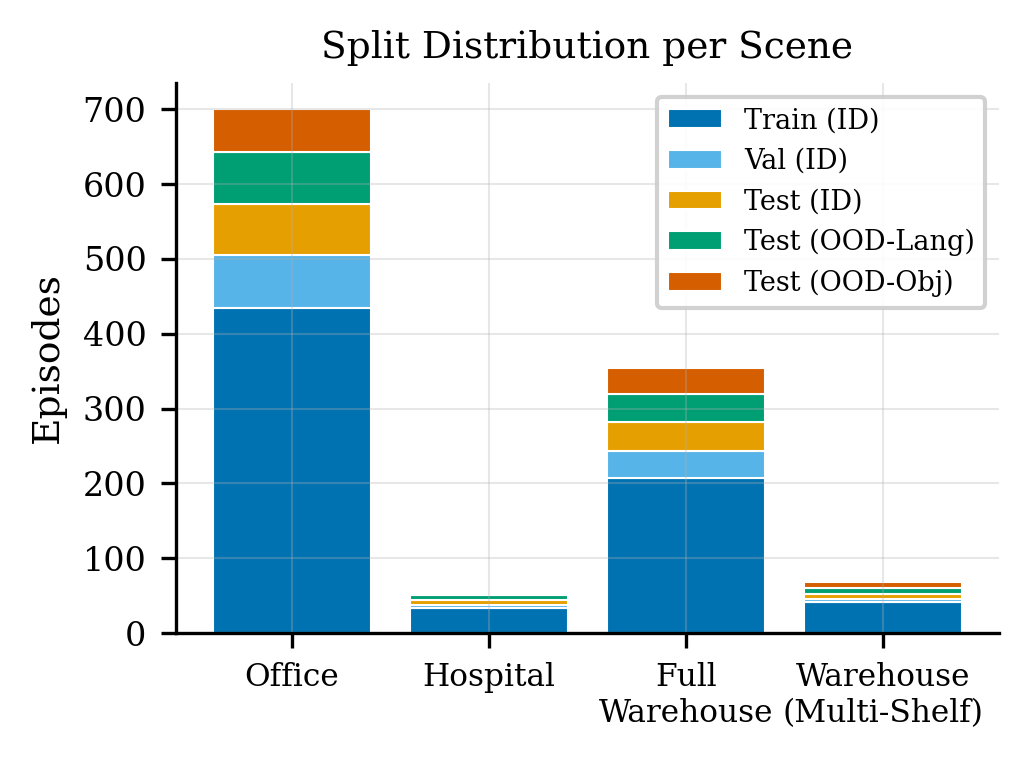}
  \caption{Episode split distribution per scene.
           Train-ID dominates each scene (60\%), with four equal-weight
           evaluation splits covering in-distribution and OOD conditions.}
  \label{fig:splits}
\end{figure}

Fig.~\ref{fig:splits} shows the scene-stratified split breakdown.

\subsection{Object Category Distribution}

\begin{figure}[t]
  \centering
  \includegraphics[width=\columnwidth]{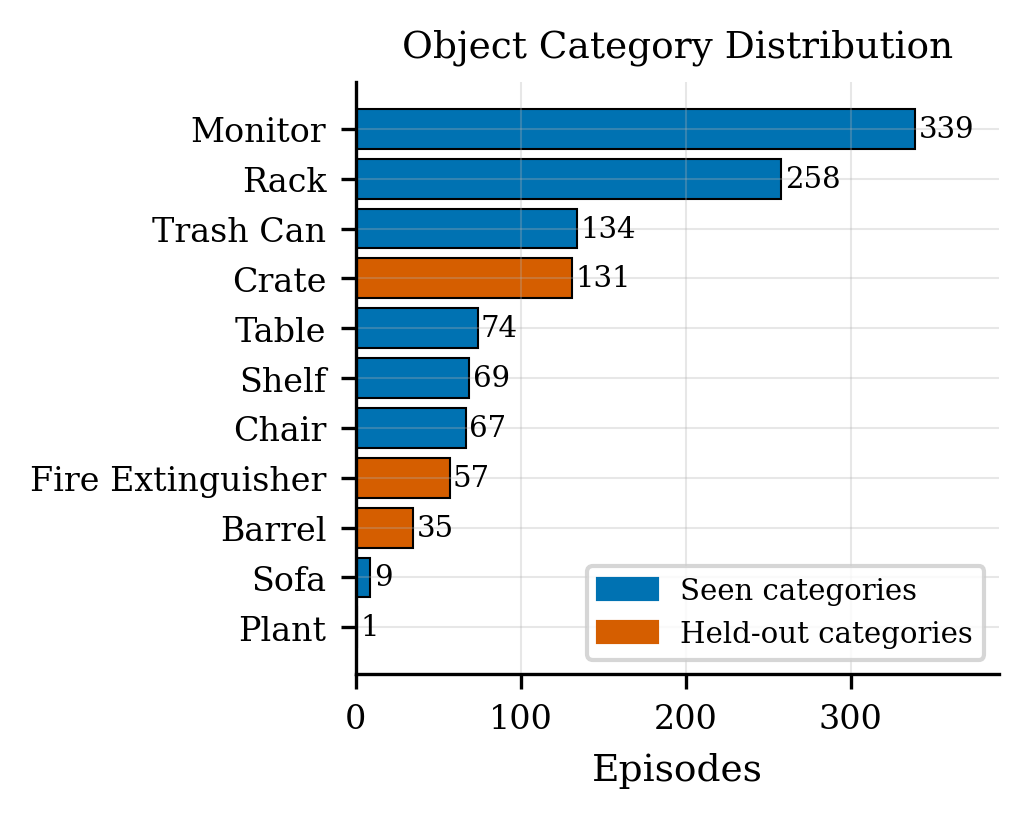}
  \caption{Object category distribution across 1\,174 episodes.
           Blue bars: seen training categories.
           Red bars: held-out OOD-object categories.
           Monitor leads with 339 episodes (office scene); rack contributes
           258 episodes from the warehouse scenes.}
  \label{fig:categories}
\end{figure}

Fig.~\ref{fig:categories} shows the category histogram.
Office-scene objects (monitor, trash\_can, table, chair) dominate the
current snapshot because the office budget is furthest along.
Held-out categories (fire\_extinguisher, whiteboard, barrel, crate) appear
exclusively in OOD-object splits.

\subsection{Trajectory Statistics}

\begin{figure}[t]
  \centering
  \includegraphics[width=\columnwidth]{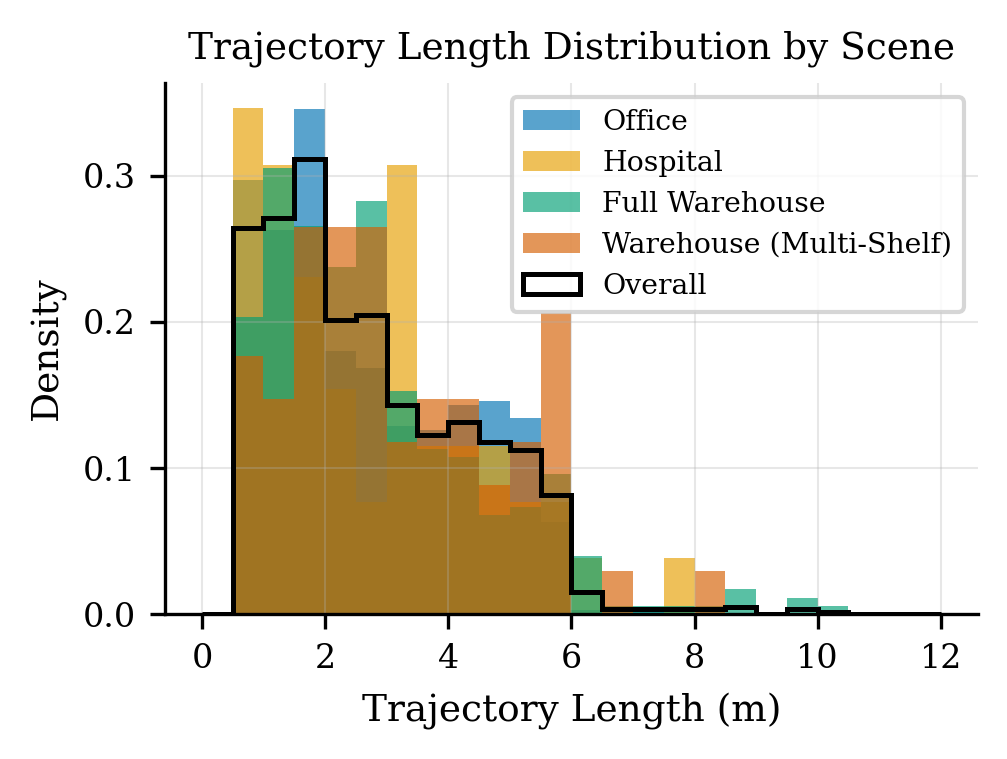}
  \caption{Trajectory length distributions per scene.
           The overall distribution (black step curve) peaks near 0.5--2\,m
           (near-tier) with a long right tail up to $\sim$10\,m (mid/far tiers),
           driven by the three spawn tiers and varying scene geometries.}
  \label{fig:tl_hist}
\end{figure}

Fig.~\ref{fig:tl_hist} shows the trajectory-length distribution.
The distribution is heavily right-skewed: most episodes (near and mid
tiers, 54.5\% and 44.3\% respectively) produce trajectories of 0.5--5\,m,
with a long tail extending to $\sim$10\,m driven by far-spawned and
warehouse episodes.
Table~\ref{tab:rollout_stats} reports per-split rollout statistics.

\begin{table}[t]
\caption{Rollout Statistics per Split (mean / median)}
\label{tab:rollout_stats}
\centering
\renewcommand{\arraystretch}{1.15}
\resizebox{\columnwidth}{!}{%
\begin{tabular}{@{}lcrrrr@{}}
\toprule
\textbf{Split} & \textbf{N} & \textbf{NE (m)} & \textbf{TL mean} & \textbf{TL med.} & \textbf{Steps mean / med.} \\
\midrule
train\_id              & 716  & 0.967 & 2.70 & 2.35 & 195.9\,/\,173 \\
val\_id                & 114  & 0.969 & 2.93 & 2.70 & 211.1\,/\,197 \\
test\_id               & 121  & 0.967 & 2.85 & 2.34 & 210.9\,/\,172 \\
test\_paraphrase\_ood  & 122  & 0.965 & 2.87 & 2.39 & 223.3\,/\,195 \\
test\_ood\_obj         & 101  & 0.969 & 2.54 & 2.24 & 186.0\,/\,170 \\
\midrule
\textbf{Overall}       & \textbf{1174} & \textbf{0.967} & \textbf{2.74} & \textbf{2.38} & \textbf{200.9\,/\,177} \\
\bottomrule
\end{tabular}}
\end{table}

\textbf{Navigation Error (NE)} is defined as the Euclidean distance
from the robot's final pose to the target centroid.
Because all retained episodes are successful, NE is constrained below
the 1.0\,m success radius; the mean of 0.967\,m indicates the robot
typically stops very close to the boundary, which is expected given
the stop-hold criterion activating at exactly $r_{\text{success}}$.

\begin{figure}[t]
  \centering
  \includegraphics[width=\columnwidth]{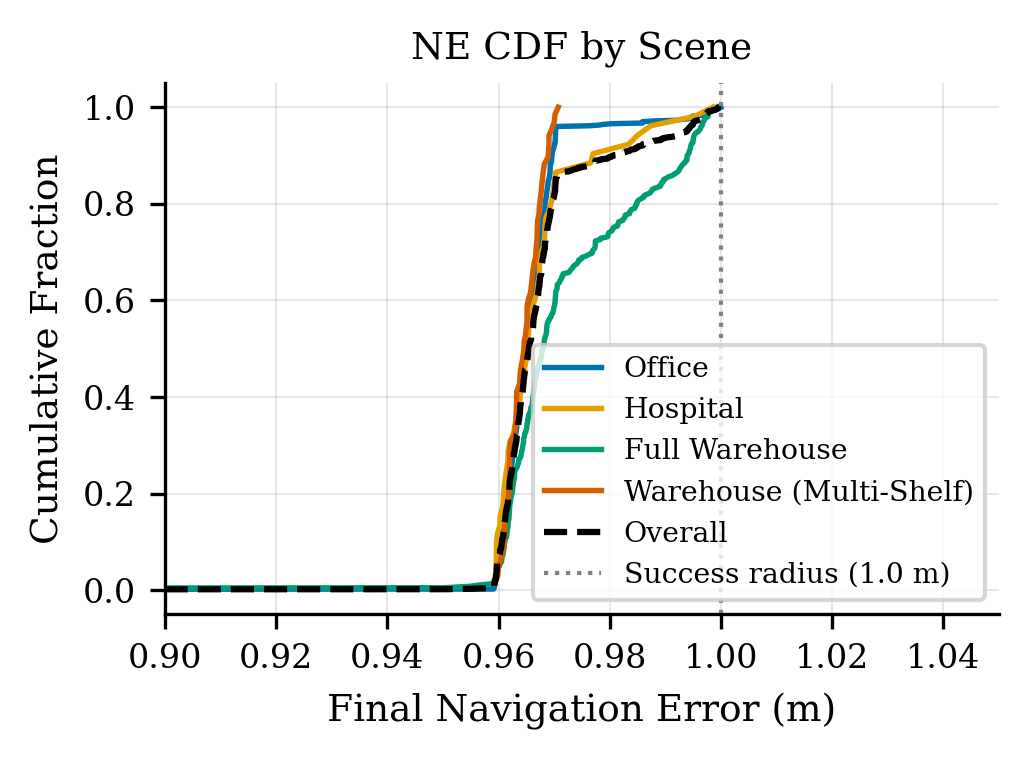}
  \caption{CDF of final navigation error (NE) per scene across 1\,174 episodes.
           All successful episodes fall within 1.0\,m (dashed vertical line).
           The tight cluster at 0.96--0.97\,m reflects the
           stop-hold controller halting at the success boundary.}
  \label{fig:ne_cdf}
\end{figure}

Fig.~\ref{fig:ne_cdf} shows the cumulative distribution of NE per scene,
confirming tight clustering just below the 1\,m threshold across all scenes.

\subsection{Episode Length Distribution}

\begin{figure}[t]
  \centering
  \includegraphics[width=\columnwidth]{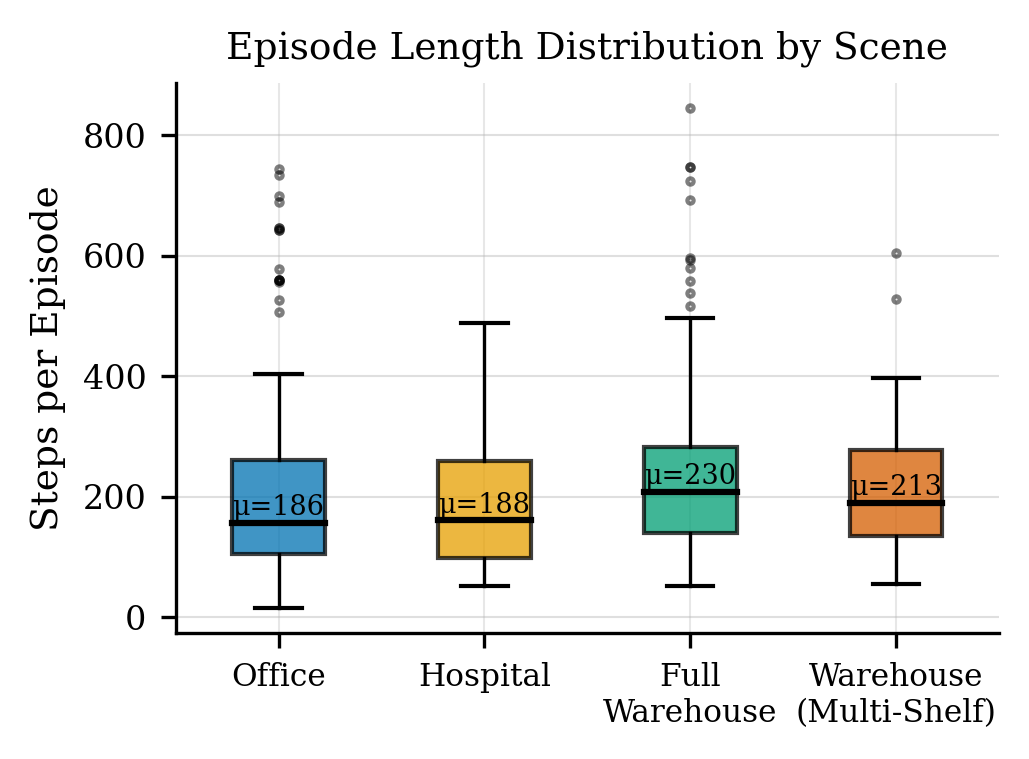}
  \caption{Episode length (steps) by scene.
           Box: interquartile range; whiskers: 1.5\texttimes{}IQR;
           dots: outliers; $\mu$: mean.
           The warehouse environments require more steps due to
           denser obstacle fields.}
  \label{fig:steps}
\end{figure}

Fig.~\ref{fig:steps} shows per-scene episode-length box plots.
Full Warehouse has the longest median episode length (207 steps)
due to its larger floor area requiring longer navigation paths,
followed by Warehouse (Multi-Shelf) at 189 steps whose cluttered
shelf geometry adds maneuvering overhead.
The Office and Hospital scenes are more open with median lengths of
156 and 160 steps respectively.

\subsection{Spawn Distance and Trajectory Length}

\begin{figure}[t]
  \centering
  \includegraphics[width=\columnwidth]{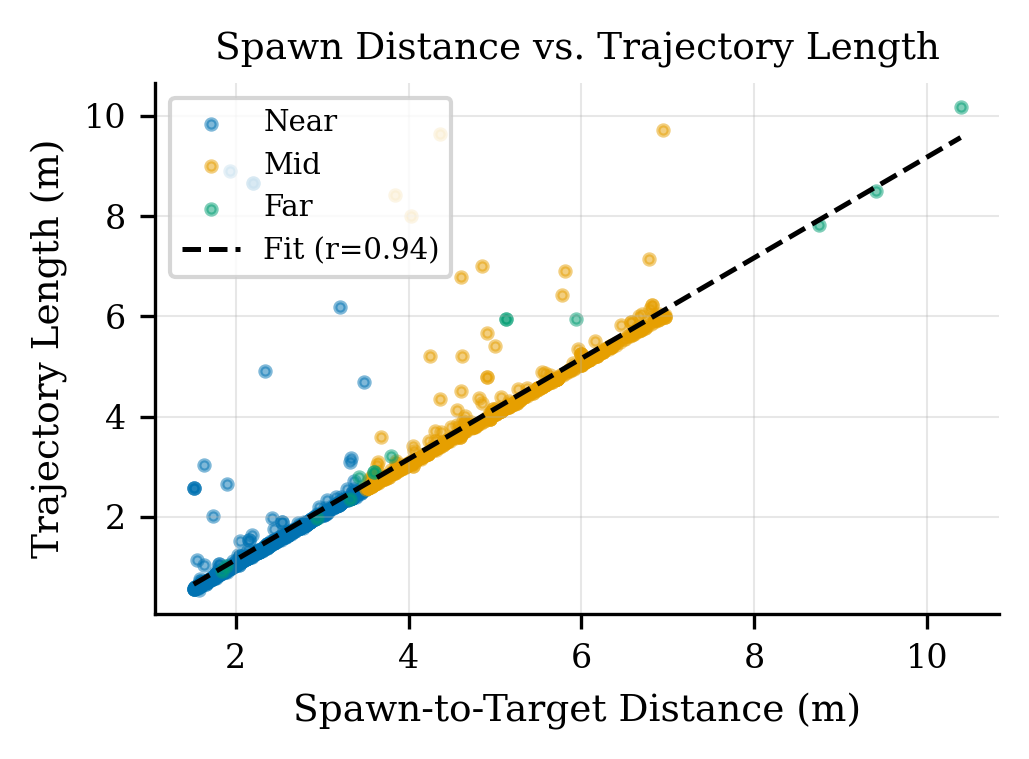}
  \caption{Spawn distance vs.\ trajectory length, coloured by spawn tier.
           A linear fit (dashed) has slope $\approx$1.00\,m/m and
           Pearson $r = 0.94$, reflecting near-direct proportionality
           between spawn distance and travel distance.}
  \label{fig:scatter}
\end{figure}

Fig.~\ref{fig:scatter} shows a strong correlation between spawn distance
and trajectory length ($r = 0.94$, 95\% CI: $[0.93, 0.95]$, $p < 0.001$,
$N = 1\,174$, slope $\approx 1.00$\,m/m).
The near-unit slope is consistent with the expert following a near-straight
path toward the target — a result of the proportional controller's design.
Rather than evidence of "difficulty diversity," this high $r$ confirms
\emph{expert navigation efficiency}: the robot travels approximately
as far as required by the spawn distance with minimal detour.
This implies that near/mid/far tiers produce systematically different
episode lengths, which is the intended training-difficulty signal.

\begin{figure}[t]
  \centering
  \includegraphics[width=\columnwidth]{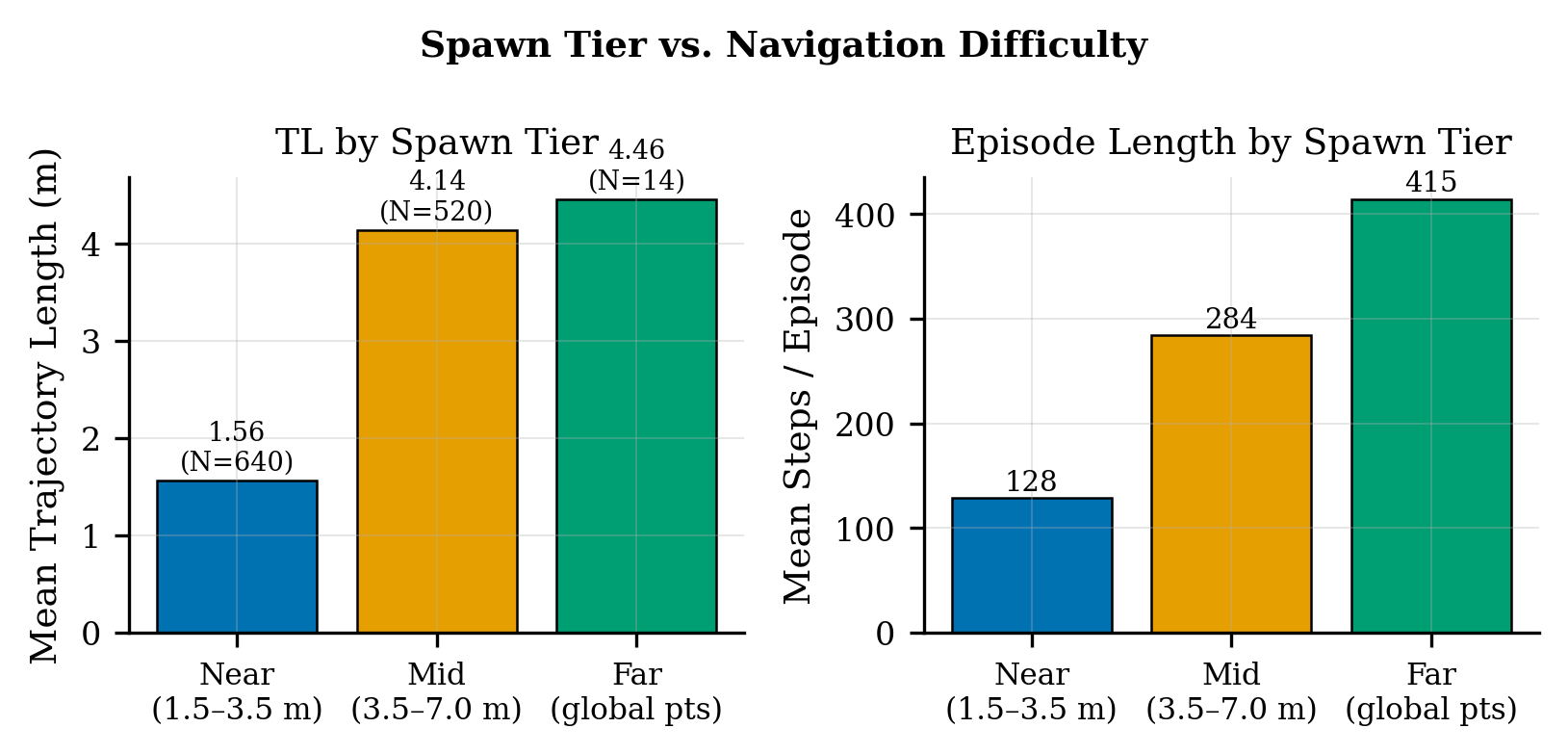}
  \caption{Mean trajectory length and episode-length steps by spawn tier.
           Far-tier episodes are $2.9\times$ longer in TL and $3.2\times$
           longer in steps than near-tier, confirming the tiers produce
           meaningfully different navigation challenges.
           Far-tier N=14 is low; this will improve as global spawn-point
           coverage expands.}
  \label{fig:tier_stats}
\end{figure}

Table~\ref{tab:tier_stats} and Fig.~\ref{fig:tier_stats} quantify this:
far-tier episodes are on average $2.9\times$ longer than near-tier
(4.46\,m vs.\ 1.56\,m mean TL) and take $3.2\times$ more steps
(415 vs.\ 129), confirming the tiers produce measurably different
navigation challenges even under the current expert.

\begin{table}[t]
\caption{Rollout Statistics by Spawn Tier}
\label{tab:tier_stats}
\centering
\renewcommand{\arraystretch}{1.15}
\resizebox{\columnwidth}{!}{%
\begin{tabular}{@{}lrrrr@{}}
\toprule
\textbf{Tier} & \textbf{N} & \textbf{Mean NE (m)} & \textbf{Mean TL (m)} & \textbf{Mean Steps} \\
\midrule
Near (1.5--3.5\,m) & 640 & 0.967 & 1.56 & 128.5 \\
Mid  (3.5--7.0\,m) & 520 & 0.967 & 4.14 & 284.3 \\
Far  (global pts)  &  14 & 0.970 & 4.46 & 414.6 \\
\midrule
\textbf{All}       & 1174 & 0.967 & 2.74 & 200.9 \\
\bottomrule
\end{tabular}}
\end{table}

\subsection{Language Template Coverage}

\begin{figure}[t]
  \centering
  \includegraphics[width=\columnwidth]{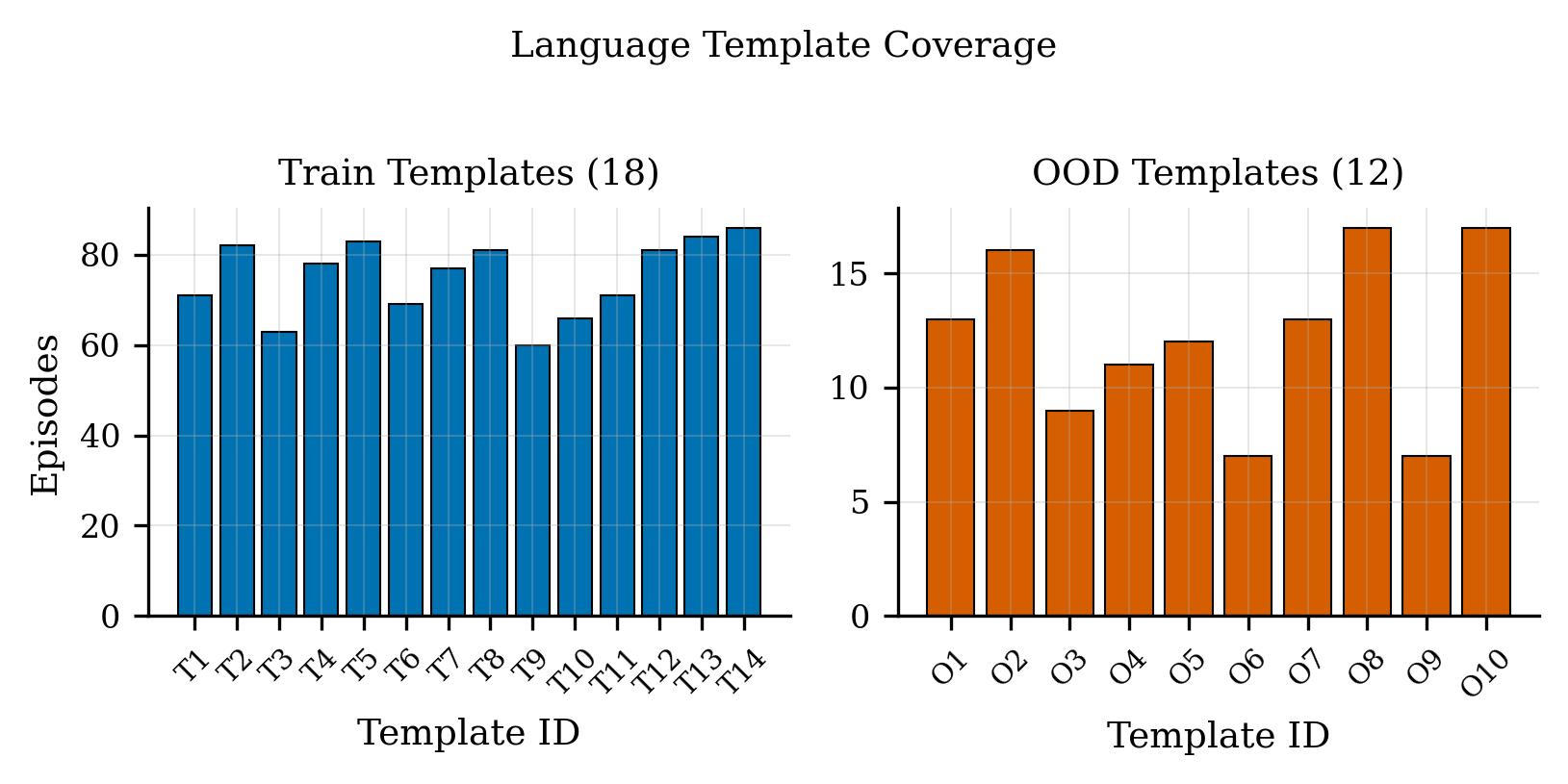}
  \caption{Episode count per language template.
           Train templates (T1--T18, left) and paraphrase-OOD templates (O1--O12, right).
           Color-slot templates (T6--T8, T16, O4, O10) have zero instances
           because all targets carry \texttt{color=unknown} in this release.}
  \label{fig:templates}
\end{figure}

Fig.~\ref{fig:templates} shows per-template episode counts.
Non-color training templates are uniformly covered (within sampling noise);
color-slot templates (T6, T7, T8, T16) have exactly zero instances because
all targets carry \texttt{color=unknown} in this release.
Paraphrase-OOD templates appear only in \texttt{test\_paraphrase\_ood} splits.

\section{Intended Use and Baselines}
\label{sec:use}

\subsection{Intended Use Cases}

MiniVLA-Nav v1 is designed to support:

\begin{itemize}
  \item \textbf{Imitation learning / behavior cloning}: continuous or tokenized
        action regression from $(o_t, \ell) \mapsto a_t$.
  \item \textbf{VLA fine-tuning}: language-conditioned navigation policies
        fine-tuned from large pretrained VLMs.
  \item \textbf{OOD generalization research}: systematic study of how
        well models transfer to unseen instruction phrasings
        (\texttt{test\_paraphrase\_ood}) and unseen object categories
        (\texttt{test\_ood\_obj}).
  \item \textbf{Sim-to-real transfer}: data collected with a real Nova Carter
        CAD model can be used to bootstrap real-robot policies.
\end{itemize}

\subsection{Suggested Evaluation Metrics}

\begin{itemize}
  \item \textbf{Success Rate (SR)}: fraction of test episodes where the
        model brings the robot within 1.0\,m of the goal and stops.
  \item \textbf{Navigation Error (NE)}: mean Euclidean distance from
        terminal pose to goal centroid.
  \item \textbf{Oracle Progress (OP)}: fraction of spawning distance
        covered toward the goal.
  \item \textbf{Collision Rate}: fraction of episodes terminated by
        stall detection.
\end{itemize}

\subsection{Baseline Expectations and Evaluation Notes}

Because all retained episodes are expert demonstrations (SR = 100\%),
a naive \emph{open-loop playback} baseline achieves SR $\approx$ 0
due to compounding errors from the first mis-prediction.
A natural evaluation ladder:

\begin{enumerate}
  \item \textbf{BC baseline}: ResNet encoder + LSTM over instruction tokens
        regressing continuous $(v,\omega)$.
        We expect SR on \texttt{test\_id} to be substantially below the expert
        due to compounding errors, with further drops on OOD splits.
  \item \textbf{Language-ablation baseline}: same architecture with a randomised
        or incorrect instruction.
        This baseline isolates how much SR is attributable to visual
        homing vs.\ genuine instruction following.
        A large gap between the language-conditioned and language-ablated
        SR validates that the dataset exercises language grounding.
  \item \textbf{VLA fine-tuning}: fine-tuning a pretrained VLM on the tokenized
        action stream.
\end{enumerate}

Full training results for all three baselines are reserved for the companion
training paper.

\section{Reproducibility and Data Availability}
\label{sec:repro}

\subsection{Generation Reproducibility}

All stochasticity in the generation pipeline is seeded through a
single top-level random seed (\texttt{seed = 42}) passed to Python's
\texttt{random.Random} and NumPy's random routines.
The generator script records the Git commit hash and Isaac Sim version
in every \texttt{dataset\_meta.json}, ensuring bit-for-bit reproducibility
of the dataset given the same simulator installation.

\subsection{Data Availability}

The dataset is publicly available on HuggingFace at
\url{https://huggingface.co/datasets/alibustami/miniVLA-Nav}.
Episodes can be iterated with standard Python filesystem APIs;
the directory structure is self-documenting as described in
Table~\ref{tab:episode_struct}.

\subsection{Ethics and Data Use}

MiniVLA-Nav v1 is entirely synthetic simulation data with no personally
identifiable information, human subjects, or sensitive content.
The Nova Carter robot USD model and the Isaac Sim environment assets
are used under the NVIDIA Omniverse License Agreement.

\section{Limitations and Future Work}
\label{sec:limits}

\noindent\textbf{Far-tier underrepresentation.}
Only 14 of 1\,174 episodes (1.2\%) are far-tier because the curated global
spawn-point files cover only a subset of floor positions per scene.
Running an expanded spawn-point survey for the warehouse scenes and rerunning
with \texttt{--resume} will correct this imbalance.

\noindent\textbf{Color annotation.}
All targets carry \texttt{color = "unknown"} because the Isaac Sim assets
do not expose material-color attributes through a standard USD API.
As a result, all color-slot templates are suppressed in the current release,
effectively reducing the training template pool from 18 to 13 active templates
and the OOD pool from 12 to 10.
Future work will add mesh-based dominant-color extraction to unlock the full
template diversity.

\noindent\textbf{Expert quality and selection bias.}
The proportional controller is near-optimal in open-plan spaces but
struggles with sharp corners and narrow doorways, leading to substantially
lower attempt-to-success rates in hospital and warehouse scenes.
This introduces a selection bias: the collected episodes disproportionately
represent unobstructed approach paths.
A path-planning expert (e.g., RRT* on a precomputed occupancy grid) would
broaden coverage and yield more naturalistic avoidance trajectories.

\noindent\textbf{Category imbalance.}
Three categories (monitor 28.9\%, rack 22.0\%, crate 11.2\%) account
for 62.1\% of episodes.
This imbalance arises because the Office scene (700 episodes, monitor-dense)
completed first.
When training on the current snapshot, a model that learns a category-specific
visual homing strategy may score higher than one that generalises across
categories.
Future collection will reduce this skew through proportional per-scene
sampling, and the language-ablation baseline
(§\ref{sec:use}) will help diagnose whether SR gains stem from visual
homing or instruction following.

\noindent\textbf{Obstacle avoidance parameters.}
The foreground crop for obstacle detection (rows 55--95\%, cols 25--75\%)
was chosen to correspond to the robot's physical collision zone at typical
navigation speeds (0.3--0.7\,m/s).
The 5th-percentile depth threshold (0.25\,m) and avoidance velocity cap
were hand-tuned; a systematic parameter search may improve warehouse-scene
SR.

\noindent\textbf{Sim-to-real domain gap.}
Three key gaps affect transfer from MiniVLA-Nav v1 to a real Nova Carter:
(1)~\emph{visual appearance} — RTX-rendered RGB differs from real camera
output in texture sharpness and lighting; (2)~\emph{depth accuracy} —
ideal simulation depth lacks the quantization noise and reflectance
artifacts of the physical Hawk stereo camera; and
(3)~\emph{floor dynamics} — PhysX rigid-body simulation does not model
real floor surface variation, wheel slip, or vibration.
Domain randomization and real-to-sim adaptation are necessary next steps
before deploying a policy trained solely on MiniVLA-Nav v1.

\noindent\textbf{Static scenes.}
All objects are static; future versions should include dynamic
obstacles (moving persons, mobile robots) to train collision avoidance
more robustly.

\noindent\textbf{Single-camera modality.}
Only the front-right stereo camera is used.
Incorporating 360\textdegree\ panoramic inputs or lidar could support
richer environmental understanding.

\section{Conclusion}
\label{sec:conclusion}

We presented MiniVLA-Nav v1, a multi-scene simulation dataset for
language-conditioned wheeled robot navigation targeting the LCOA task.
With 1\,174 episodes across four photorealistic Isaac Sim scenes,
12 object categories, 30 language templates, and three spawn-distance tiers,
the dataset provides multi-modal per-step observations and structured OOD
evaluation splits for systematic VLA research.

Three design properties are validated by the 1,174-episode snapshot:
(1) the expert navigates efficiently ($r = 0.94$ between spawn distance and
TL, slope $\approx$1.00\,m/m), producing a clean difficulty signal for
the three spawn tiers;
(2) the stop-hold criterion yields tight terminal NE clustering
(mean 0.967\,m $<$ 1.0\,m success radius, std $\approx$0.001\,m);
and (3) warehouse scenes produce $2.2\times$ longer episodes (median
189--207 steps) than office/hospital scenes (156--160 steps), confirming
that scene geometry drives meaningful difficulty diversity.

The companion training paper will report BC, language-ablated, and VLA
fine-tuning baselines on all five evaluation splits.
Near-term roadmap: expand far-tier spawn coverage, deploy a path-planning
expert for higher warehouse coverage, and extract mesh-based color
annotations to activate the remaining color-slot templates.

\bibliographystyle{IEEEtran}
\bibliography{minivla_nav_v1}

\end{document}